\documentclass[sigconf, screen ]{acmart}

\AtBeginDocument{%
  }
\usepackage{bm}
\usepackage[table]{xcolor}
\usepackage{multirow}

\setcopyright{acmlicensed}
\copyrightyear{2018}
\acmYear{2026}
\acmDOI{XXXXXXX.XXXXXXX}
\acmISBN{978-1-4503-XXXX-X/2018/06}

\begin{document}

\title{GraphFusion3D: Dynamic Graph Attention Convolution with Adaptive Cross-Modal Transformer for 3D Object Detection}


\author{Md Sohag Mia}
\email{shuvo2018@nuist.edu.cn}
\authornotemark[1]
\affiliation{%
  \institution{Bangladesh Universtiy of Engineering and Technology}
  \city{Dhaka}
  \country{Bangladesh}
}

\author{Md Nahid Hasan}
\email{nahid.hasan.bondhan@gmail.com}
\affiliation{%
  \institution{Bangladesh Universtiy of Engineering and Technology}
  \city{Dhaka}
  \country{Bangladesh}}

\author{Muhammad Abdullah Adnan}
\email{adnan@cse.buet.ac.bd}
\affiliation{%
  \institution{Bangladesh Universtiy of Engineering and Technology}
  \city{Dhaka}
  \country{Bangladesh}
}

\renewcommand{\shortauthors}{Trovato et al.}

\begin{abstract}
  Despite significant progress in 3D object detection, point clouds remain challenging due to sparse data, incomplete structures, and limited semantic information. Capturing contextual relationships between distant objects presents additional difficulties. To address these challenges, we propose GraphFusion3D, a unified framework combining multi-modal fusion with advanced feature learning. Our approach introduces the Adaptive Cross-Modal Transformer (ACMT), which adaptively integrates image features into point representations to enrich both geometric and semantic information. For proposal refinement, we introduce the Graph Reasoning Module (GRM), a novel mechanism that models neighborhood relationships to simultaneously capture local geometric structures and global semantic context. The module employs multi-scale graph attention to dynamically weight both spatial proximity and feature similarity between proposals. We further employ a cascade decoder that progressively refines detections through multi-stage predictions. Extensive experiments on SUN RGB-D (70.6\% AP\(_{25}\) and 51.2\% AP\(_{50}\)) and ScanNetV2 (75.1\% AP\(_{25}\) and 60.8\% AP\(_{50}\)), demonstrate a substantial performance improvement over existing approaches.
\end{abstract}





\received{20 February 2007}
\received[revised]{12 March 2009}
\received[accepted]{5 June 2009}

\maketitle

\section{Introduction}
\label{sec:intro}

%----------------------------------------------------------------------------

% ==== FIG 1
\begin{figure*} [t]
  \centering
  \includegraphics[width=7.1in]{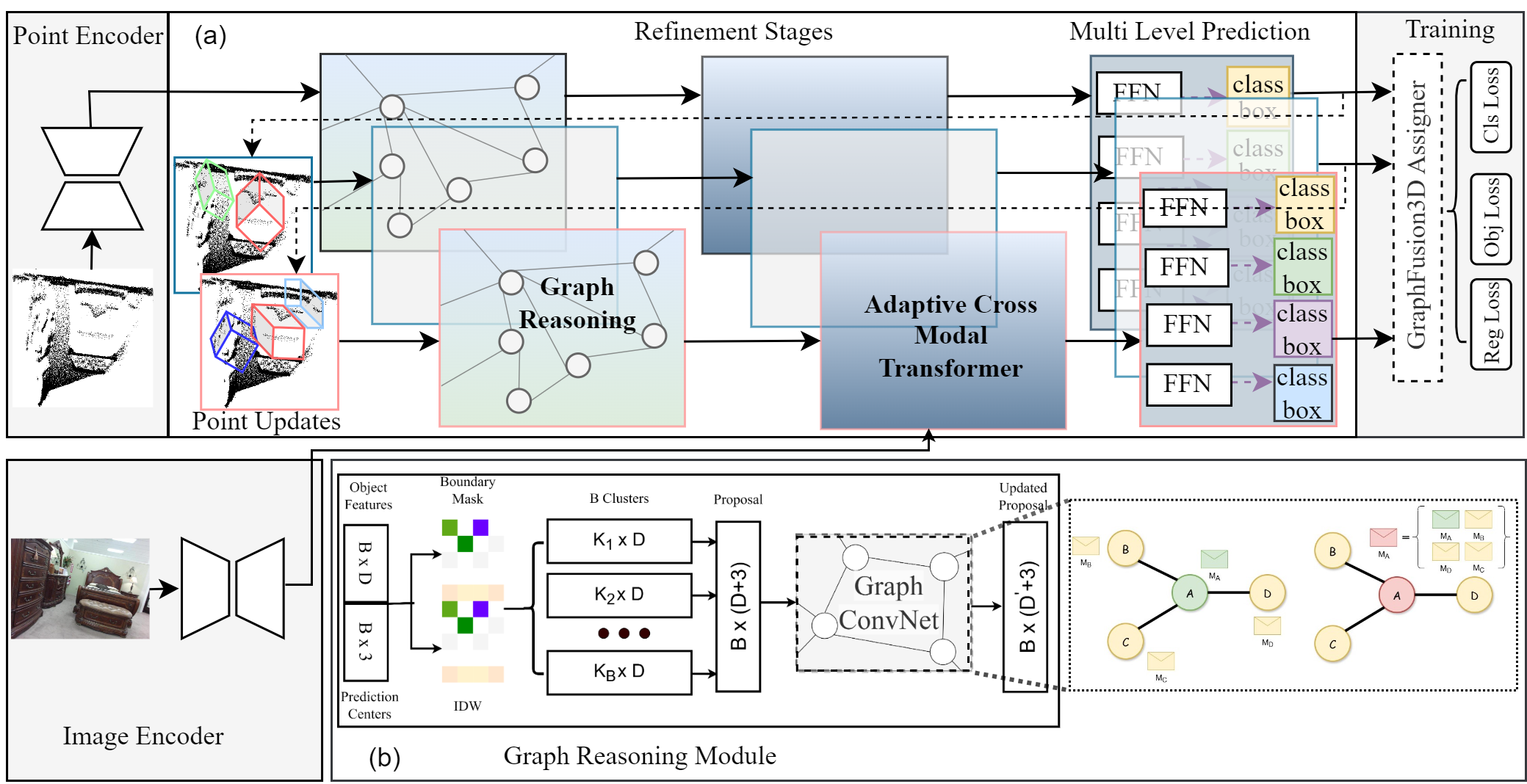}
  \caption{\textbf{Our proposed GraphFusion3D architecture}. The framework processes point clouds and RGB images through four key stages: (1) feature extraction via point and image backbone networks, (2) contextual refinement using the Graph Reasoning module, (3) multi-modal fusion via Adaptive Cross-Modal Transformer, and (4) progressive Cascade Refinement Decoding. Final detections are generated after multi-stage refinement.}
\label{graphfusion3d}
\end{figure*}

%----------------------------------------------------------------------------

Recently, 3D object detection from point clouds has become increasingly popular. As a core technique for 3D scene understanding, it is being utilized in many fields, such as domestic robotics, autonomous vehicles, and augmented reality. Accurate 3D detection is crucial for these applications because it enables machines to interact safely and intelligently with real-world environments, which are inherently three-dimensional. For example, autonomous vehicles rely on precise 3D perception to detect pedestrians, other vehicles, and road structures in complex, dynamic conditions. Similarly, service robots operating in indoor environments must be able to understand object shapes and positions to grasp or navigate around them reliably. These detection methods can be broadly categorized into two groups: 3D convolution-based approaches and voting-based approaches. Voting-based methods \cite{10} cluster features via proposals but scale poorly when dealing with large point clouds due to high computational costs and the need for multiple passes over the data. Sparse 3D convolutions use voxels for efficient, scalable processing without losing density, enabling models to handle high-resolution point clouds with fewer resources. While most 3D models rely on point clouds, integrating multi-modal data \cite{imvotenet}, such as combining RGB images with LiDAR or depth scans, enhances detection accuracy by supplementing sparse geometry with rich semantic cues like texture and color. Some approaches depend solely on image data \cite{im1,im2}. While images offer rich color, texture, and semantic details, their lack of depth data limits their ability to predict precise 3D bounding boxes for 3D detection tasks. Some approaches depend solely on point clouds \cite{10,pointnet,PointNet++,7}, but point clouds are often irregular, sparse, and unordered, with varying point densities that may result in missing details, especially for small or distant objects. Small objects often get removed from the point cloud data, which can lead to blind regions and inaccurate scene representation. Information provided by the two different types of datasets is not similar to each other; rather, they complement each other, making multi-modal fusion a promising direction for robust perception. Recent research works focus on this limitation by utilizing both types of datasets \cite{imvotenet,frustum,DeMF,EPNet++}, combining geometric and semantic information to produce more accurate and complete 3D detections. These models can utilize both features, which in this case improves accuracy and robustness against occlusion and missing data. In order to capture the image features, ImVoteNet \cite{imvotenet} utilizes a faster RCNN model, and EPNet++ \cite{EPNet++} utilizes an encoder-decoder model with convolutional blocks.

However, existing multi-modal approaches often rely on fixed or early fusion strategies that fail to dynamically balance geometric cues from sparse point clouds with rich semantic information (color, texture) from images, particularly when one modality is unreliable due to occlusion, noise, or limited views. This leads to suboptimal fusion in complex indoor scenes with varying object scales and partial observations.
To address this gap, our proposed method advances 3D object detection through several key innovations. We employ Deform-DETR to extract discriminative image features from RGB inputs while developing a novel Graph Reasoning Module (GRM) that enriches proposal features with neighborhood context. The framework incorporates a customized Deformable Feature Fusion module that dynamically aligns and combines multi-modal features from both point clouds and images. Further refinement is achieved through a cascade decoding process that progressively improves detection quality across multiple stages. Together, these components demonstrate the effectiveness of hybrid architectures for high-performance 3D perception in complex environments. 
Our overall work can be summarized as follows:

\begin{itemize}
    \item  We introduce a novel Graph Reasoning for Proposal Context, a multi‐scale learnable graph module that captures semantic and geometric relations among image‐point proposals.
    
    \item We propose an Adaptive Cross-Modal Transformer (ACMT), which dynamically aligns 2D image features and 3D point features in a context-aware manner, enabling effective image–point fusion and enriching both geometric and semantic representations for robust 3D object detection.
    
    \item Finally, we design a Progressive Cascaded Refinement Decoder, which iteratively refines detection proposals in a cascaded manner, improving both accuracy and localization.
\end{itemize}

\section{Related Work}
\label{sec:related-works}

We contextualize our approach within 3D object detection research, focusing on data modalities and architectural paradigms. Current methods primarily employ either point clouds or multi-modal inputs, implemented through voting, expansion, or transformer frameworks. While point-based approaches capture precise geometry, they often lack semantic richness—a limitation addressed by multi-modal methods through RGB fusion. This section reviews these critical developments.

\subsection{Point Clouds Based}
PointNet \cite{pointnet} processes irregular point sets via T-Net transformations and MLPs, extracting global features for detection. PointNet++ \cite{PointNet++} extends the original PointNet by capturing local geometric structures through a hierarchical network that learns features at multiple scales for more robust 3D point cloud understanding. VoteNet \cite{10} demonstrates how combining deep point set learning with Hough voting effectively addresses the challenge of detecting objects directly in sparse point clouds. It shows that strong 3D detection performance can be achieved using purely geometric information, without depending on 2D image data. GroupFree \cite{Group-Free} leverages a transformer to progressively update object queries and integrate iterative predictions for refined 3D detection. 3DETR \cite{7} brought an end-to-end transformer approach to the 3D detection of point clouds. The class-aware grouping strategy together with sparse convolutional RoI pooling in CAGroup3D \cite{cagroup3d} leads to better 3D object detection results on ScanNet and SUN RGB-D benchmarks with improved accuracy and efficiency. The Vertex Relative Position Encoding (3DV-RPE) method in V-DETR \cite{v-detr} boosts 3D object detection performance through attention mechanisms that focus on spatially related points according to the locality principle. The technique obtains leading performance results for ScanNetV2 by surpassing CAGroup3D with significant improvement. SPGroup3D \cite{spgroup3d} utilizes superpoint grouping methods which integrate geometry-based voting together with feature fusion to achieve better semantic matching and precise one-stage 3D object detection for indoor environments. UniDet3D \cite{unidet3d} solves the problem of dataset-dependant 3D object detectors by bringing together multiple indoor datasets using a common label system which results in enhanced model robustness and generalization capabilities. Through its transformer architecture the system delivers notable performance improvements on six key benchmarks which proves its suitability for multiple indoor use cases. SOFW \cite{sofw} presents a unified framework which optimizes indoor 3D object detection across multiple datasets through simultaneous domain parameter optimization between shared and specific domains while maintaining model complexity at a constant level.

\subsection{Multi-Modal Based}
The ImVoteNet \cite{imvotenet} system improves VoteNet through its integration of 2D visual and 3D point cloud data, which solves missing information problems in point clouds. The system integrates two independent backbones along with specialized towers which enhance multi-modal fusion capabilities for detection. The method \cite{2} merges 3D CNN depth feature representations with 2D VGG color outputs by combining them through concatenation to perform SUN RGB-D \cite{SUNRGB} 3D detection. By integrating RGB and point clouds through transformers, TokenFusion \cite{tokenfusion} dynamically replaces low-information tokens in one modality with fused cross-modal features, boosting accuracy in geometrically or texturally ambiguous regions. EPNet++ \cite{EPNet++} model merges RGB and point cloud data at intermediate backbone layers, refining cross-modal alignment through progressive feature fusion. \cite{DeMF} Employs dual pipelines for image and point cloud data, using a DeMF module to fuse object-level visual information into point features, but skips the feature refinement process. MTC-RCNN \cite{MTC-RCNN} cascades a 3D detection framework with intermediate 2D segmentation, leveraging pixel-level semantics to improve proposal refinement. FCAF3D \cite{fcaf3d} implements full convolutional anchor-free detection through point-based proposal parametrization but faces challenges with irregular geometries because noisy coarse proposal points do not align properly with ground truth centers. TR3D \cite{tr3d} introduces a lightweight fully-convolutional 3D object detector that enhances sparse CNNs for better accuracy and efficiency, and its TR3D+FF version uses early fusion of RGB and point cloud features to deliver strong multimodal performance on SUN RGB-D. However, such early/fixed fusion strategies do not adaptively weigh modalities based on local context or reliability, limiting robustness when image views are sparse or points are noisy. Our ACMT with Cross-Modal Gating (CMG) explicitly addresses this by dynamically balancing image semantics and point geometry per query.
\section{THE PROPOSED METHOD}
\label{sec:method}

In this section, we present our proposed framework for 3D object detection, illustrated in Fig.~\ref{graphfusion3d}. Starting from a standard sparse 3D convolutional backbone and image feature extractor, we first generate initial proposals via set aggregation. These proposals are then enhanced through three integrated components: (1) a Graph-Aware Voting Mechanism that enriches proposal features by leveraging neighborhood context; (2) an Adaptive Cross-Modal Transformer (ACMT) that effectively integrates multi-modal data from images and point clouds; and (3) a Progressive Cascaded Refinement Decoder; Below, we detail each component in sequence.

%-------------------------------------------
% =======
% FIG. 02
% =======
%\begin{figure}[t]
%  \centering
%  \includegraphics[width=3.3in]{sec/figure/GAV.png}\\
%  \caption{The architecture of Graph-Aware Voting for Contextual Proposal Refinement.}
%\label{GAV}
%\end{figure}
%-------------------------------------------

%--------------------------------------------------------------------------------------------
%--------------------------------------------------------------------------------------------

\subsection{Graph Reasoning Module}

We propose a novel Graph Reasoning Module (GRM) that models both geometric and semantic relationships among proposals via multi-scale graph attention convolutional networks (depicted in Fig.~\ref{graphfusion3d}(b)).  
The GRM explicitly encodes contextual dependencies, leveraging the fact that certain spatial configurations recur in indoor layouts (e.g., chairs near tables, beds beside nightstands).  
This module integrates three key operations: boundary-aware feature aggregation, multi-scale attention-based graph convolution, and residual fusion.

The first stage performs \textit{boundary-aware feature aggregation} using conditional inverse distance weighting (IDW), ensuring that features are aggregated only from spatially relevant regions:
\begin{equation}
    f_{i}=\frac{\sum_{j \in \mathcal{N}(i)} \zeta_{i,j} w_{i,j} f_{j}}{\sum_{j \in \mathcal{N}(i)} \zeta_{i,j} w_{i,j}},
    \quad
    w_{i,j} = \exp\!\left(-\frac{\|p_i' - p_j\|}{\sigma}\right),
\label{ia}
\end{equation}
where $p_i'$ is the proposal coordinate, $f_j$ denotes neighboring point features, $\zeta_{i,j}$ is a boundary mask (the boundary mask is determined by checking whether the neighboring point lies inside the proposal's bounding box extent, preventing feature leakage across object boundaries), and $\sigma$ is set to the mean $k$-NN distance to stabilize aggregation across variable densities.  
This normalization avoids bias from dense neighborhoods and ensures spatially consistent feature refinement.

Next, we construct an attention-based graph convolution that captures both geometric and feature-level dependencies:
\begin{equation}
e_{ij}=h_\varphi(x_i, x_j, p_i - p_j),
\label{eq:gcn}
\end{equation}
where $h_\varphi$ is a nonlinear function with learnable parameters $\varphi$.  
Edge weights are computed from both feature similarity and spatial distance across three neighborhood scales $k \in \{5, 10, 20\}$:

\begin{equation}
w_{ij}^{(s)}=\exp\!\left(-\frac{\|p_i - p_j\|^2}{2\sigma_s^2}\right),
\end{equation}
where $\sigma_s$ represents the adaptive scale parameter derived from the mean distance to the $k$-th nearest neighbor, ensuring stable edge weight computation across varying point densities. These spatial weights are then combined with feature-based attention coefficients:

\begin{equation}
    \alpha_{ij}^{(s)}=\frac{\exp(\text{sim}(q_i^{(s)}, k_j^{(s)}))}{\sum_{m\in\mathcal{N}_s(i)}\exp(\text{sim}(q_i^{(s)}, k_m^{(s)}))}.
\end{equation}
where $q_i^{(s)}$ and $k_j^{(s)}$ are learnable query and key vectors derived from node features, and $\text{sim}(\cdot, \cdot)$ denotes cosine similarity. This dual-weighting mechanism dynamically balances spatial proximity and feature relevance, allowing the model to focus on semantically important relationships while respecting geometric constraints.

The aggregated representation for each scale is then:
\begin{equation}
\tilde{f}_i^{(s)}=\sum_{j\in\mathcal{N}_s(i)} \alpha_{ij}^{(s)}w_{ij}^{(s)}\psi^{(s)}(e_{ij}^{(s)}),
\end{equation}
followed by scale fusion and residual update:
\begin{equation}
f_i'' = f_i + \gamma \cdot \text{MLP}([\tilde{f}_i^{(5)};\tilde{f}_i^{(10)};\tilde{f}_i^{(20)}]),
\end{equation}
where $\gamma$ is a learnable scalar.

GRM is applied twice within the transformer decoder—once before decoding to establish a coarse relational context, and again after the first decoding stage for fine-grained relational refinement. This two-step reasoning enables both object-level and scene-level contextual understanding. GRM produces context-aware proposal relationships without relying on external segmentation, enabling seamless integration within the unified GraphFusion3D pipeline.

% \noindent\textbf{Discussion.}

%--------------------------------------------------------------------------------------------

%-------------------------------------------
% =======
% FIG. 03
% =======
\begin{figure}[t]
  \centering
  \includegraphics[width=3.3in]{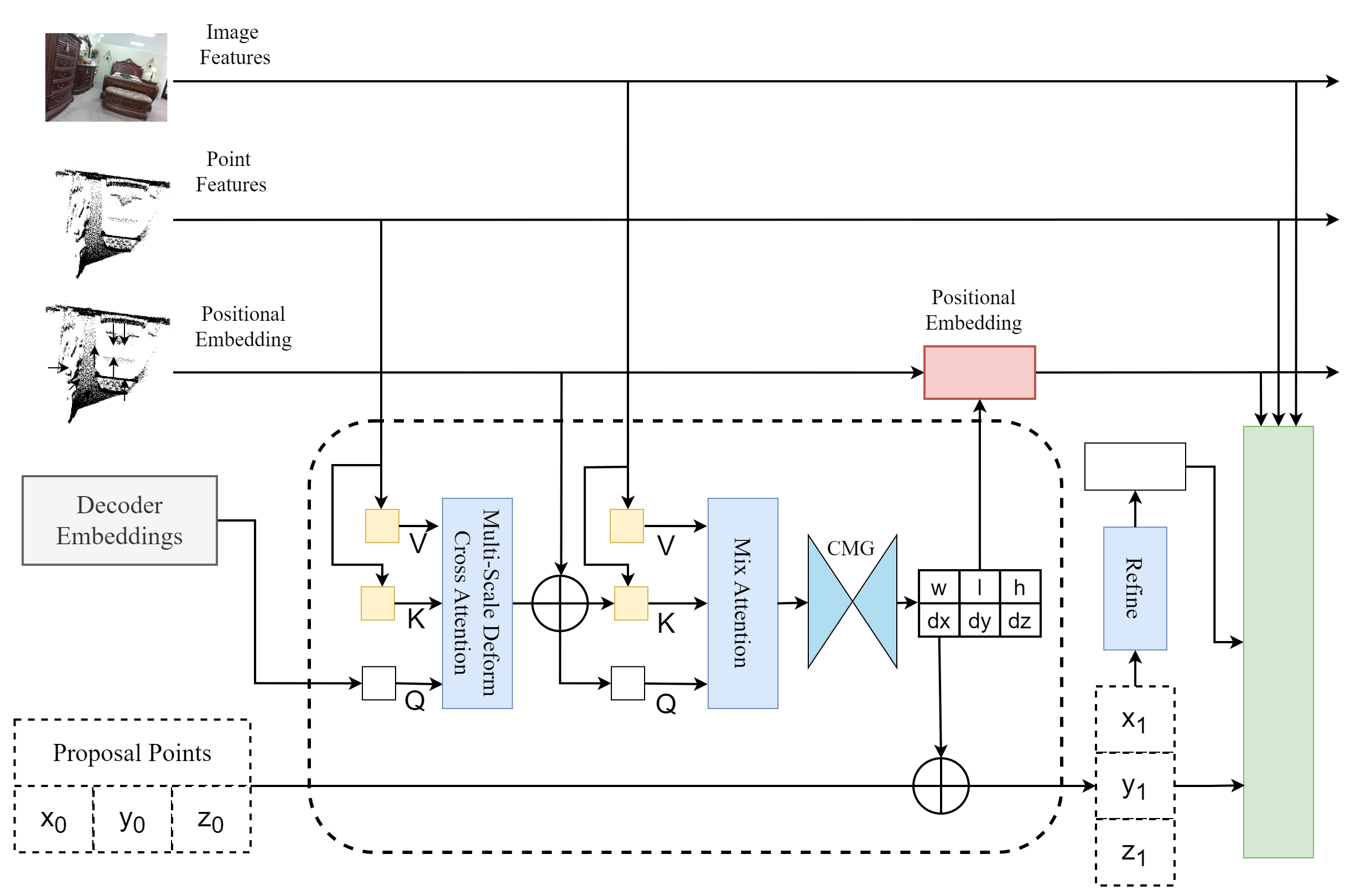}
  \caption{The architecture of the Adaptive Cross-Modal Transformer module.}
\label{acmt}
\end{figure}
%-------------------------------------------

%-------------------------------------------
% =======
% FIG. 04
% =======
\begin{figure}[t]
  \centering
  \includegraphics[width=3.3in]{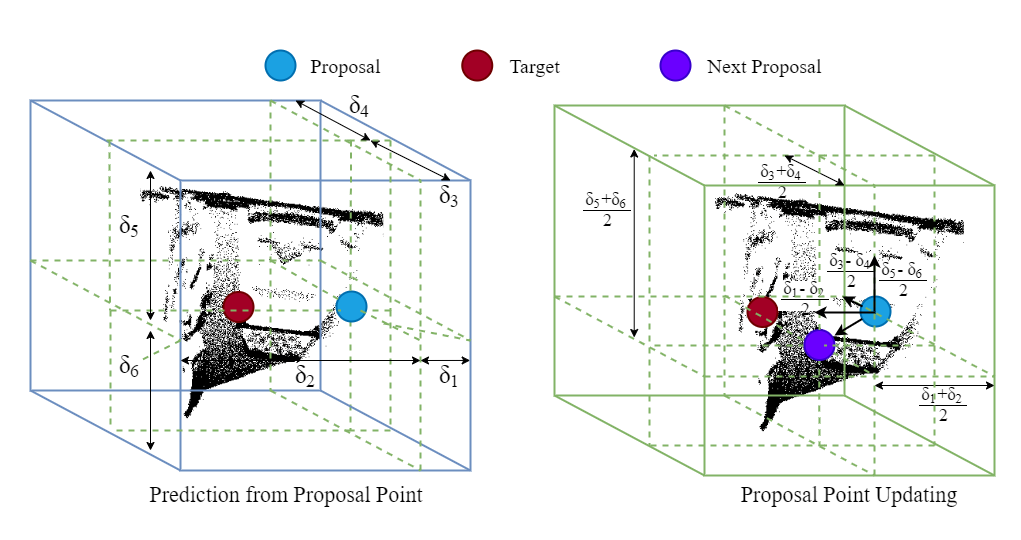}\\
  \caption{Illustration of a progressive cascaded refinement decoder module. Left: Bounding box from a single proposal point with predicted $\delta$. Right: Updating process from proposal point to bounding box center.}
\label{cascade_decoder}
\end{figure}
%-------------------------------------------

%-------------------------------------------------------------------------------------

\subsection{Adaptive Cross-Modal Transformer}

The Adaptive Cross-Modal Transformer (ACMT) operates on the relationally structured proposals from GRM, using them as geometric anchors for its cross-modal attention.
ACMT is the core fusion module in GraphFusion3D, designed to unify geometric information from 3D points and semantic cues from RGB images in a spatially consistent manner. 
ACMT alternates attention between the point-cloud and image modalities, guided by adaptive modality weighting through a \textit{Cross-Modal Gating (CMG)} mechanism. 
This design allows the network to dynamically balance 2D and 3D cues depending on object scale, visibility, and context. The ACMT module is shown in Fig.~\ref{acmt}.

\noindent\textbf{Feature Extraction.} 
We employ a ResNet backbone to extract multi-scale image features $\{x^l_i\}_{l=1}^L$ ($L=4$) and a sparse 3D convolutional encoder to obtain geometric features $x^p \in \mathbb{R}^{N \times C}$. 
The two feature streams are spatially aligned through explicit geometric projection.

Each 3D point $\mathbf{P}=[x,y,z,1]^{\top}$ in camera coordinates is projected onto the image plane using the intrinsic matrix $\mathbf{K}$ and extrinsic parameters $[\mathbf{R}|\mathbf{t}]$:
\[
\begin{bmatrix}
\tilde{u} \\ \tilde{v} \\ \tilde{w}
\end{bmatrix}
= \mathbf{K}[\mathbf{R}|\mathbf{t}]\mathbf{P},
\quad
(u, v) = \left(\frac{\tilde{u}}{\tilde{w}}, \frac{\tilde{v}}{\tilde{w}}\right).
\]
The resulting coordinates $(u,v)$ are then normalized to the range $[0,1]^2$ using the image dimensions:
\[
\text{refPoint} = \left(\frac{u}{W_0}, \frac{v}{H_0}\right),
\]
where $W_0$ and $H_0$ denote the image width and height, respectively. 
This normalization ensures consistent pixel-level alignment between projected 3D points and 2D feature maps, allowing ACMT to sample semantically relevant regions during fusion.

\noindent\textbf{Cross-Modal Fusion.} 
Given the aligned features, ACMT performs alternating attention between modalities to refine object queries $y$. 
At each layer, point features \(x^p\) provide geometric priors through cross-attention, while image embeddings \(x^i\) contribute semantic detail through deformable attention:
\begin{equation}
\begin{array}{l@{\;}l}
y &= y + \text{CrossAttn}(y, x^p), \\
y &= y + \text{MSDeformAttn}(y, x^i), \\
y &= y + \text{FFN}(y),
\end{array}
\label{eq:acmt_fusion}
\end{equation}
where $\text{CrossAttn}$ models spatial dependencies in the 3D domain and $\text{MSDeformAttn}$ selectively aggregates multi-scale image features around projected reference points. 
This formulation allows ACMT to reason jointly over geometric structure and visual appearance while maintaining computational efficiency.

\noindent\textbf{Cross-Modal Gating (CMG).} 
To adaptively balance geometric and visual cues, CMG predicts modality-specific weights conditioned on the current query context and both feature streams. 
Given query features $y$, point features $y^p$, and image features $y^i$, the gating network first concatenates them and passes the result through a multi-layer perceptron (MLP) with layer normalization and ReLU activation:
\[
[\lambda_p, \lambda_i] = \text{Softmax}\big(\text{MLP}([y; y^p; y^i])\big),
\]
where $\lambda_p, \lambda_i \in \mathbb{R}^{H}$ correspond to per-head weights for point and image modalities, respectively. 
The fused representation is then computed as:
\[
y' = \lambda_p \odot y^p + \lambda_i \odot y^i,
\]
where $\odot$ denotes element-wise weighting across attention heads. 
This fine-grained gating mechanism dynamically adjusts the contribution of each modality, emphasizing geometric structure when appearance cues are unreliable and amplifying semantic detail when visual context is strong.
ACMT integrates the gating mechanism directly within the transformer layers rather than using it as an external weighting step. This joint formulation enables context-aware reweighting at every fusion stage and stabilizes gradient flow between 2D and 3D branches. The result is a more robust cross-modal alignment capable of handling complex indoor scenes with partial observations.

%--------------------------------------------------------------------------------------------
%--------------------------------------------------------------------------------------------

\subsection{Progressive Cascaded Refinement Decoder}
Using the relational–semantic features produced by ACMT, the Progressive Cascaded Refinement Decoder updates box geometry in a multi-stage fashion.
The Progressive Cascaded Refinement Decoder iteratively improves proposal localization and feature quality through multi-stage refinement.  
Given initial proposals $(\mathbf{x}^p, \mathbf{p})$, each stage predicts box offsets $\bm{\delta} = \{\delta_1, \dots, \delta_6\}$ relative to the ground-truth box $\bm{g} = (x, y, z, w, l, h, \theta)$:
\begin{equation}
\begin{split}
\delta_1 = x + \tfrac{w}{2} - \hat{x}, \quad
\delta_2 = \hat{x} - x + \tfrac{w}{2}, \\
\delta_3 = y + \tfrac{l}{2} - \hat{y}, \quad
\delta_4 = \hat{y} - y + \tfrac{l}{2}, \\
\delta_5 = z + \tfrac{h}{2} - \hat{z}, \quad
\delta_6 = \hat{z} - z + \tfrac{h}{2}.
\end{split}
\end{equation}
To bring proposals closer to ground truth centers, the decoder updates the 3D coordinates at each stage as:
\[
\hat{x}' = \hat{x} + \tfrac{\delta_1 - \delta_2}{2}, \;
\hat{y}' = \hat{y} + \tfrac{\delta_3 - \delta_4}{2}, \;
\hat{z}' = \hat{z} + \tfrac{\delta_5 - \delta_6}{2}.
\]

Centerness supervision encourages points closer to the true object center. Let the centerness target \( \hat{c}^* \) be defined based on the normalized regression targets \( \delta^* \) as:
\[
\hat{c}^* = \prod_{i=1}^{3} \frac{\min(\delta^*_{2i-1}, \delta^*_{2i})}{\max(\delta^*_{2i-1}, \delta^*_{2i})}.
\]
A point is assigned positive if it lies within the ground-truth spatial region, reducing noise from distant samples. Updating process is shown in Fig.~\ref{cascade_decoder}.

\noindent\textbf{GraphFusion3D Assigner.} Existing assigners \cite{fcaf3d,tr3d} often fail with thin objects or sparse point distributions due to strict boundary constraints. Our GraphFusion3D Assigner overcomes this by combining boundary-aware matching with cascade threshold scheduling. It progressively decreases the positive assignment threshold across decoder stages while considering points both inside and near bounding boxes, ensuring abundant training samples initially and high-quality positives later for challenging objects.

\noindent\textbf{Training Objective.}
The final optimization combines classification, regression, and centerness losses:
\[
\mathcal{L} = \lambda_{\text{cls}}\mathcal{L}_{\text{cls}} + 
\lambda_{\text{reg}}\mathcal{L}_{\text{reg}}^{\text{RotatedIoU3D}} + 
\lambda_{\text{ctr}}\mathcal{L}_{\text{ctr}},
\]
with $\lambda_{\text{cls}}{=}1.0$, $\lambda_{\text{reg}}{=}2.0$, $\lambda_{\text{ctr}}{=}1.0$.  
This balanced loss design stabilizes training and ensures accurate 3D localization and class prediction across refinement stages.

%xxxxxxxxxxxxxxxxxxxxxxxxxxxxxxxxxxxxxxxxxxxxxxxxxxxxxxxxxxxxxxxxxxxxxxxxxxxxxxxxxxxxxxxxxxxxxxxxxx

%------------------------------------------------------------------------------------------------
% =======
% TABLE. 02
% =======
\begin{table*}[t]
\label{tab:table2}
%\resizebox{\textwidth}{!}{%
\setlength{\tabcolsep}{5.5pt} 
\begin{tabular}{@{}l | c |  llllllllll  ll@{}}
\hline

Methods &  Modal & Bed & Table & Sofa & Chair & Toilet & Desk & Dresser & Stand & Shelf & Tub & AP\(_{25}\) & AP\(_{50}\) \\ 

\hline

VoteNet \cite{10} &   P & 83.0 & 47.3 & 64.0 & 75.3 & 90.1 & 22.0 & 29.8 & 62.2 & 28.8 & 74.4 & 57.7 &  \\

3DETR \cite{7}  &  P  &  81.8 & 50.0 & 58.3 & 68 & 90.3 & 28.7 & 28.6 & 56.6 & 27.5 & 77.6 & 59.1  & 32.7 \\

GroupFree \cite{Group-Free}  &   P &  87.8   &  53.8   &  70.0   &  79.4   &  91.1   &  32.6    &   36.0   &   66.7   & 32.5     &   80.0   & 63.0 & 45.2 \\

FCAF3D  \cite{fcaf3d}   &   P  &  87.6   &  53.6   &  70.7   &   81.5  &  92.3   &  37.1    &  39.8    & 70.2     &   34.5   &  80.1    & 64.2 & 48.9 \\

HGNet \cite{chen2020hierarchical}  &    P &   84.5  &  51.6   &  65.7   &  75.2   &  91.1   & 34.3     &   37.6   &   61.7   &  35.7    &   78.0   & 61.6 &   \\

TR3D  \cite{tr3d}   & P  &  88.7   &  58.5   &   71.8  &   82.8  &  92.9   &   41.7   &   44.8   &    73.1  &  36.3    &  78.4    &   67.1 & 50.4  \\

SPGroup3D  \cite{spgroup3d}  &  P & 89.6  &  55.9   &   74.9  &  81.0   &  90.3   &   38.6   &   37.8   &   66.4   &  38.1    &   78.6   & 65.4 & 47.1 \\

CAGroup3D \cite{cagroup3d}  & P  &  89.1   &   59.6  &   73.6  &  83.9   &  91.7   &   40.6   &  40.5    &  72.9    &   35.4   & 82.6    &  66.8   &  50.2    \\

V-DETR  \cite{v-detr}   & P  &     &     &     &     &     &      &      &      &      &      &  67.5   &  50.4  \\

Uni3DETR \cite{uni3detr}   &  P &   &     &     &     &     &      &      &      &      &      &   67.0   &  50.3   \\

MMTC \cite{MTC-RCNN}  &   P+I  &   86.3  &   54.4  &  68.2   &   79.7  &  92.8   &   29.7   &  47.2    &   69.5   &   46.3   & 78.8     &   65.3  &  48.6   \\

TR3D+FF \cite{tr3d}  & P+I &     &     &     &     &     &      &      &      &      &      & 69.4 & 53.4 \\

EPNet\(++\)  \cite{EPNet++} &   P+I & 89.1 & 51.3 & 71.9 & 80.2 & 92.4 & 32.5 & 45.2 & 67.4 & 47.1 & 76.3 & 65.3 &   \\

TokenFusion \cite{tokenfusion}  &  P+I &     &     &     &     &     &      &      &      &      &      & 64.9 & 48.3 \\

ImVoteNet \cite{imvotenet} & P+I  & 87.6 & 51.1 & 70.7 & 76.7 & 90.5 & 28.7 & 41.4 & 69.9 & 41.3 & 75.9 & 64.4 & 43.3 \\

\hline
\rowcolor{gray!15}
GraphFusion3D  & P &  86.1   &  53.7   &  69.3   &   80.3  &  92.7   &  36.1    &  39.3    & 70.2     &   34.0   &  78.2   & 64.0 & 47.1 \\
\rowcolor{gray!15}
GraphFusion3D  & P+I  & 92.5 & 62.1 & 78.8 & 86.5 & 92.8 & 48.9 & 45.7 & 82.0 & 31.7 & 84.0 & 70.6 & 51.2 \\

\hline
\end{tabular}
%}
\caption{State-of-the-art results comparison of each object, AP\(_{25}\) and AP\(_{50}\) on SUN RGB-D. P= Point only input, and P+I= Point + RGB inputs.}
\label{sunrgbd}
\end{table*}

%--------------------------------------------------------------------------------------------
%--------------------------------------------------------------------------------------------

%--------------------------------------------------------------------------------------------

% =======
% TABLE. 01
% =======

\begin{table}[t]
%\label{tab:table1}
\centering
\setlength{\tabcolsep}{12.5pt} 
\begin{tabular}{l c c c}
\hline
\textbf{Methods} & Modal & \textbf{AP\(_{25}\)} & \textbf{AP\(_{50}\)} \\ 
\hline

VoteNet & P &       58.6 & 33.5 \\

3DETR   & P &       65.0  & 47.0  \\ 

FCAF3D  &  P &      71.5   &  57.3 \\ 

SPGroup3D & P &     74.3 & 59.6 \\

Uni3DETR  &  P &    71.7  & 58.3  \\ 

CAGroup3D &  P &    75.1 & 61.3 \\ 

UniDet3D & P  &     77.5 & 66.1 \\ 

TR3D   &  P &       72.9 & 59.3 \\ 

V-DETR  &  P &      77.4 &  65.0 \\ 

\hline
\rowcolor{gray!15}
GraphFusion3D  & P & 72.8 &  57.9  \\ 

\rowcolor{gray!15}
GraphFusion3D  & P+I & 75.1 &  60.8  \\ 
\hline
\end{tabular}
\caption{ Quantitative evaluation on ScanNetV2. P= Point only input, and P+I= Point + RGB inputs.}
\label{scannet}
\end{table}

%--------------------------------------------------------------------------------------------

% =======
% TABLE. 02
% =======

\begin{table}[t]
\centering
\setlength{\tabcolsep}{1pt}
\begin{tabular}{lccccc}
\toprule
Backbones           & Point backbone            & Image backbone         & AP\(_{25}\) & AP\(_{50}\) \\
\midrule
CNNs                & MinkResNet34      & ResNet     & 70.6  &  51.2 \\

\rowcolor{gray!15}
Transformers        & Swin3D       & Swin-T      & 71.8  & 52.6 \\
\bottomrule

\end{tabular}
\caption{Effect of different backbones on SUNRGB-D dataset.}
\label{tab:backbone_comparison}
\end{table}

%--------------------------------------------------------------------------------------------

%--------------------------------------------------------------------------------------------
% =======
% TABLE. 04
% =======

\begin{table}[t]
\centering
\setlength{\tabcolsep}{8.5pt}
\begin{tabular}{lcccc}
\hline
Module & Params (M) &  FPS & AP\(_{25}\) & AP\(_{50}\) \\
\hline
Baseline            & 103.7 &  16.39 & 64.0  & 47.1  \\
+ ACMT              & 124.6 &  12.76 & 67.3 & 49.2\\
+ CRD               & 124.6 & 10.18 & 68.4  & 49.4\\
\rowcolor{gray!15}
+ GRM               & 126.2 &  10.01 & 70.6  & 51.2 \\
\hline
\end{tabular}
\caption{Performance impact of different modifications on SUN RGBD and module-wise runtime and parameter breakdown of GraphFusion3D.}
\label{tab:module_efficiency}
\end{table}

%--------------------------------------------------------------------------------------------

%--------------------------------------------------------------------------------------------
%--------------------------------------------------------------------------------------------

\section{EXPERIMENTS}
\label{sec:exp}

%--------------------------------------------------------------------------------------------
%--------------------------------------------------------------------------------------------
\subsection{Datasets and Evaluation Metrics}

\noindent\textbf{Datasets.} 
We conduct experiments on two standard indoor benchmarks: SUN RGB-D~\cite{SUNRGB} and ScanNetv2~\cite{scannet}.  
\textbf{(a) SUN RGB-D} consists of 10,335 RGB-D images annotated with oriented 3D bounding boxes across 37 categories. 
Following VoteNet, we evaluate on the 10 most frequent categories using the official train/val split (5,285 / 5,050 samples).  
\textbf{(b) ScanNetv2} contains 1,513 reconstructed 3D indoor scenes with per-point semantic and instance annotations across 18 object categories, divided into 1,201 training and 312 validation scans.

\noindent\textbf{Evaluation Metric.} For all datasets, we use mean average precision (AP) at IoU thresholds of 0.25 and 0.5, the same as VoteNet. To ensure reliability, We train each model five times and evaluate each trained model five times independently. For comparison, we report the best results obtained across all runs. 

\subsection{Implementation Details}
Our implementation follows standard 3D detection pipelines \cite{fcaf3d}. We uniformly sample \(N = 100{,}000\) points per scene with a voxel size of 0.01 and employ a single set aggregation for proposal generation. Each proposal is iteratively refined through bounding-box regression and classification branches. The model is trained for 12 epochs using the AdamW optimizer with an initial learning rate of \(1\times10^{-4}\), decayed by 0.1 at the 8th and 11th epochs. We apply a weight decay of 0.01 and use a batch size of 4 on a single NVIDIA RTX 4090 GPU. Losses are jointly applied at proposal and final prediction stages to stabilize training.

%--------------------------------------------------------------------------------------------
%--------------------------------------------------------------------------------------------
\subsection{Comparisons with State-of‐the-Art}

\begin{figure*}[t]
  \centering
  \includegraphics[width=7in ]{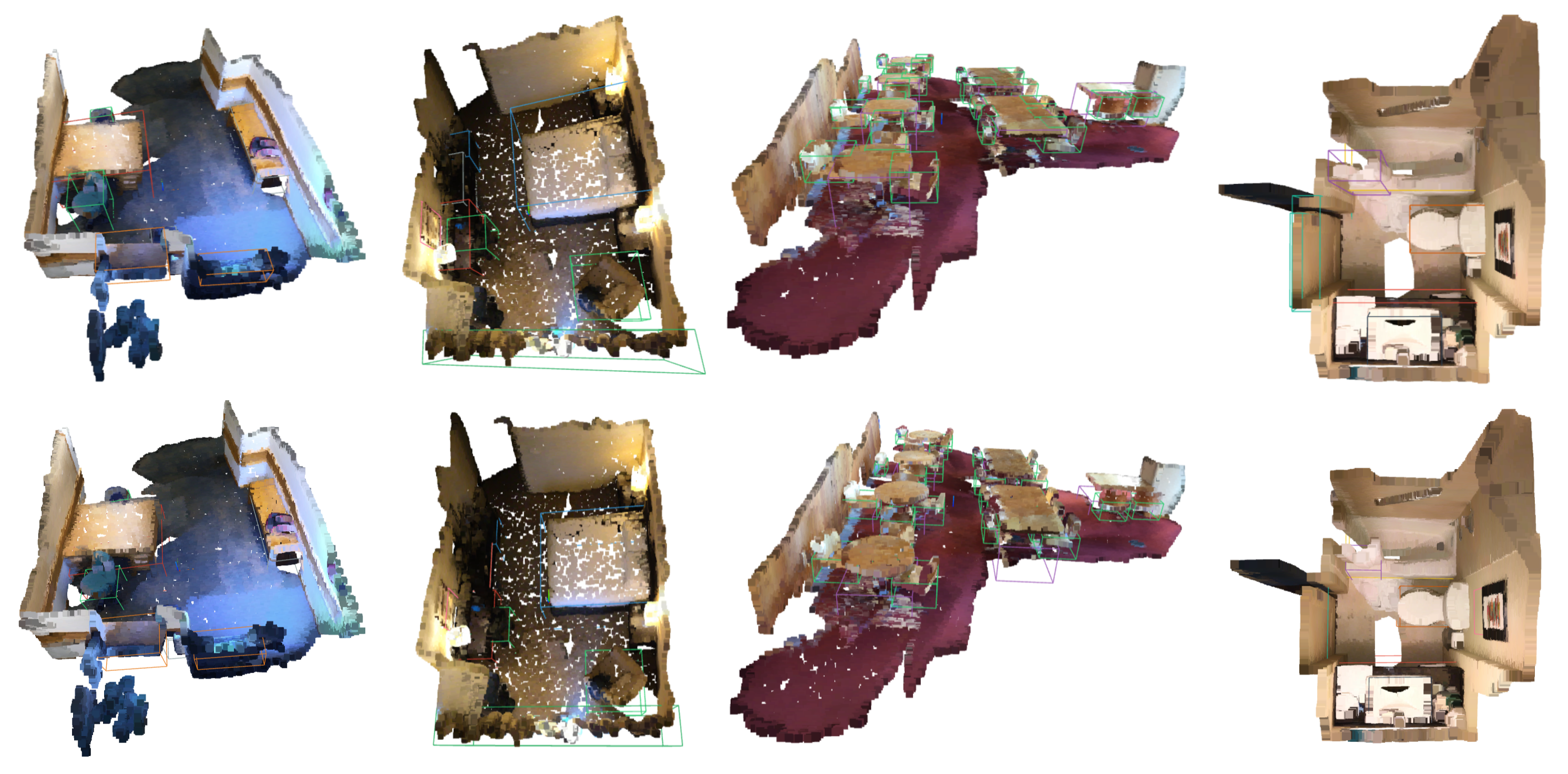}
  \caption{Qualitative prediction results on the ScanNetV2. The first row displays the ground-truth data, while the second row presents the detection results from our GraphFusion3D.}
  \label{predict-scannet}
\end{figure*}

\begin{figure}[t]
  \centering
  \includegraphics[width=3in]{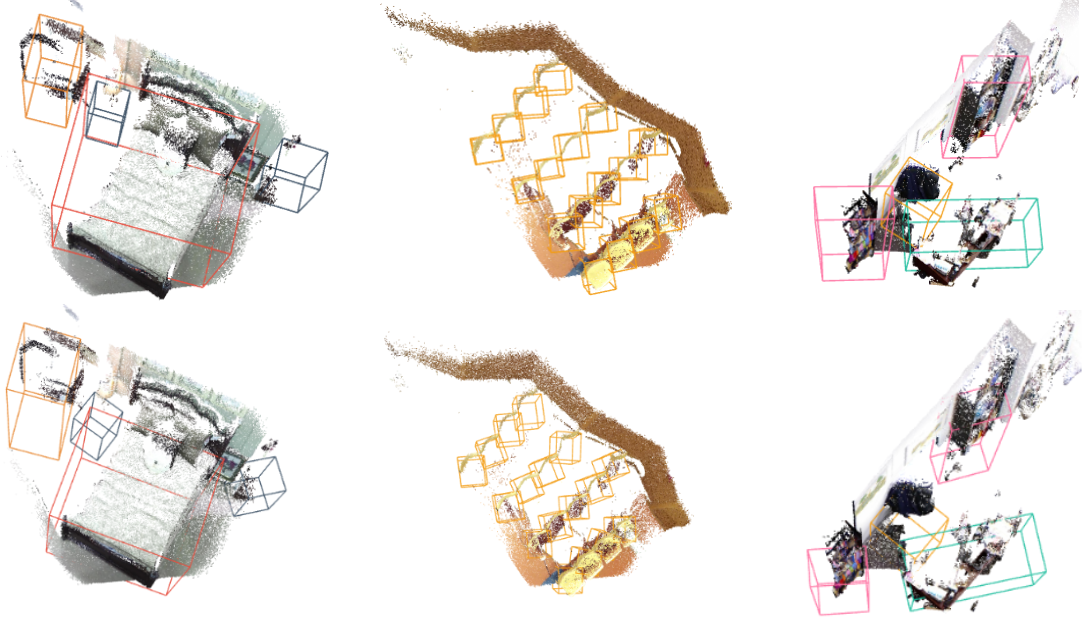}
  \caption{Qualitative prediction results on the SUN RGB-D. The first row displays the ground-truth data, while the second row presents the detection results from our GraphFusion3D.}
  \label{predict-sunrgbd}
\end{figure}

We evaluate GraphFusion3D on two different datasets. On SUN RGB-D benchmark, with results summarized in Table~\ref{sunrgbd}. Our model achieves 70.6 AP\(_{25}\) and 51.2 AP\(_{50}\), surpassing recent state-of-the-art 3D detection frameworks such as V-DETR (67.5 AP\(_{25}\)), Uni3DETR (67.0 AP\(_{25}\)), TR3D+FF (69.4 AP\(_{25}\)), and CAGroup3D (66.8 AP\(_{25}\)). GraphFusion3D attains notable category-level gains for challenging classes like chairs, sofas, and beds, highlighting its effectiveness in modeling both detailed geometric structures and global scene context. These improvements arise from the synergy between the Graph Reasoning and Adaptive Cross-Modal Transformer modules, which together enhance spatial reasoning and image-point alignment. On ScanNetV2, our multimodal model (P+I) achieves 75.1 AP\(_{25}\) and 60.8 AP\(_{50}\) (Table~\ref{scannet}). While these results demonstrate the effectiveness of our fusion approach, they do not reach the current state-of-the-art for this dataset. This is primarily because our method is specifically optimized for multimodal fusion, and we used only 10 multi-view images per scan during training—significantly fewer than the 200+ views available in the original dataset. This limitation in visual data coverage constrained our model's ability to fully exploit the complementary RGB information available in ScanNetV2. Nevertheless, the consistent improvement over our point-only variant (72.8 AP\(_{25}\)) validates the benefits of our cross-modal fusion design.
Fig.~\ref{predict-scannet} and Fig.~\ref{predict-sunrgbd} present qualitative detection results on ScanNetV2 and SUN RGB-D, respectively.

%------------------------------------------------------------------------------

\begin{table}[t]
\centering
\setlength{\tabcolsep}{9.5pt} 
\begin{tabular}{lccc}
\hline
Setting & $k$ & AP\(_{25}\) & AP\(_{50}\) \\
\hline
No Graph Reasoning        & --    & 68.4 & 49.4 \\
Single-Scale (Local)      & 5     & 68.6 & 49.5 \\
Single-Scale (Mid-range)  & 10    & 68.9 & 50.0 \\
Single-Scale (Global)     & 20    & 69.8 & 50.4 \\
\rowcolor{gray!15}
Multi-Scale (k=5,10,20)   & All   & \textbf{70.6} & \textbf{51.2} \\
\hline
\end{tabular}
\caption{Ablation study on the Graph Reasoning module with different neighborhood scales $k$ on SUN RGB-D. Multi-scale fusion shows the strongest performance by combining complementary local, mid-range, and global contexts.}
\label{tab:gav_scales}
\end{table}

%-------------------------------------------------------

\begin{table}[t]
\centering
\setlength{\tabcolsep}{4pt}
\begin{tabular}{l | cc | cc}
\toprule
\multirow{2}{*}{Decoder Stages} & \multicolumn{2}{c}{SUNRGB-D} & \multicolumn{2}{| c}{ScanNetV2} \\
\cmidrule(lr){2-3} \cmidrule(lr){4-5}
 & AP\(_{25}\) & AP\(_{50}\) & AP\(_{25}\) & AP\(_{50}\) \\
\midrule
Single-stage (Baseline)   & 69.5 & 50.9 & 73.9 & 59.6 \\
Two-stage                 & 70.4 & 51.1 & 74.8 & 60.5 \\

\rowcolor{gray!15}
Full Cascade (Three-stage) & 70.6 & 51.2 & 75.1 & 60.8 \\
\bottomrule
\end{tabular}
\caption{Effect of different cascaded stages.}
\label{cascade_stages}
\end{table}

%---------------------------------------------------------------

\begin{table}[t]
\centering
\setlength{\tabcolsep}{4pt}
\begin{tabular}{l | cc |cc}
\toprule
\multirow{2}{*}{Fusion Variant} & \multicolumn{2}{c}{SUNRGB-D} & \multicolumn{2}{| c}{ScanNetV2} \\
\cmidrule(lr){2-3} \cmidrule(lr){4-5}
 & AP\(_{25}\) & AP\(_{50}\) & AP\(_{25}\) & AP\(_{50}\) \\
\midrule
w/o Cross-Modal Gating        & 70.3 & 50.6 & 74.7 & 59.9 \\

\rowcolor{gray!15}
w/ Cross-Modal Gating         & 70.6 & 51.2 & 75.1 & 60.8 \\

\bottomrule
\end{tabular}
\caption{Ablation study of the Adaptive Cross-Modal Transformer.}
\label{acmt_ablation}
\end{table}

%--------------------------------------------------------------------------------------------

\subsection{Efficiency and Complexity Analysis}

We analyze the efficiency of our proposed modules in Table~\ref{tab:module_efficiency}. The baseline model achieves 64.0\% AP\(_{25}\) with 103.7M parameters. Incorporating the Adaptive Cross-Modal Transformer (ACMT) increases parameters to 124.6M and improves AP\(_{25}\) to 67.3\%. Adding the Cascade Refinement Decoder (CRD) maintains the same parameter count while further boosting AP\(_{25}\) to 68.4\%. The full model with the Graph Reasoning Module (GRM) uses 126.2M parameters and achieves 70.6\% AP\(_{25}\), demonstrating that significant performance gains are attained with only a moderate increase in model size. Despite the added complexity, our design maintains a competitive inference speed of 10.01 FPS.

%--------------------------------------------------------------------------------------------
%--------------------------------------------------------------------------------------------

% =======
% FIG. 06
% =======

% =======
% FIG. 03
% =======

\subsection{Ablation Studies}
We perform extensive ablation studies on the SUN RGB-D and ScanNetV2 datasets to isolate the contribution of each core module in GraphFusion3D. All experiments use the same training protocol and evaluation setup for fairness.

\noindent\textbf{Impact of Adaptive Cross-Modal Transformer.}  The Adaptive Cross-Modal Transformer (ACMT) strengthens image–point fusion through the proposed Cross-Modal Gating (CMG) mechanism, which dynamically regulates the contribution of geometric and visual features. Starting from the point-only baseline of 64.0 AP\(_{25}\), adding ACMT (without GRM and CRD)  significantly improves performance to 67.3 AP\(_{25}\) on SUN RGB-D (Table~\ref{tab:module_efficiency}. CMG learns adaptive modality weights $(\lambda_p, \lambda_i)$ conditioned on each query representation, allowing the model to emphasize point-based geometry when visual cues are unreliable (e.g., occlusion or depth noise) and to rely more on image features when appearance information is strong. Without CMG, the baseline ACMT achieves 70.3~AP\(_{25}\) and 50.6~AP\(_{50}\) on SUNRGB-D and 74.7~AP\(_{25}\) and 59.9~AP\(_{50}\) on ScanNetV2, whereas incorporating CMG improves alignment and robustness to 70.6~AP\(_{25}\) and 51.2~AP\(_{50}\) on SUNRGB-D and 75.1~AP\(_{25}\) and 60.8~AP\(_{50}\) on ScanNetV2 (Table~\ref{acmt_ablation}).

%-----------------------------------------------------------------------------------------

\noindent\textbf{Backbone Choice.}  
We also examine the impact of different backbone configurations (Table~\ref{tab:backbone_comparison}). Standard CNN backbones (MinkResNet34 \cite{Choy_2019_CVPR} for point clouds and ResNet \cite{He_2016_CVPR} for images) yield strong performance with 70.6 AP\(_{25}\) and 51.2 AP\(_{50}\). Replacing them with transformer-based architectures (Swin3D \cite{10901941} and Swin-T \cite{Liu_2021_ICCV}) further improves performance to 71.8 AP\(_{25}\) and 52.6 AP\(_{50}\), demonstrating enhanced context modeling and long-range reasoning. These results confirm that our fusion design generalizes effectively and benefits from the superior global dependency capture of transformers.

\noindent\textbf{Impact of Cascaded Refinement.} As shown in Table~\ref{cascade_stages}, our progressive cascade decoder consistently improves performance across both datasets. On SUN RGB-D, performance increases from 69.5 AP\(_{25}\)/50.9 AP\(_{50}\) (single-stage) to 70.4/51.1 (two-stage) and finally 70.6/51.2 (three-stage). Similarly, ScanNetV2 improves from 73.9/59.6 to 75.1/60.8. Each stage enhances box localization through iterative center refinement and feature updating, with diminishing returns after the second stage.

\noindent\textbf{Graph Reasoning Module.} The proposed Graph Reasoning Module module contributes significant improvement, delivering a +2.2~AP\(_{25}\) gain through explicit reasoning over spatial and semantic relationships between proposals. Unlike standard convolutional aggregation, GRM constructs a learnable graph where each proposal interacts with its neighbors across multiple receptive-field scales $k=\{5,10,20\}$. At each scale, edge affinities are computed by jointly considering spatial distance (via a Gaussian kernel) and feature similarity (via cosine attention), allowing the model to dynamically balance geometric proximity and semantic consistency.  As shown in Table~\ref{tab:gav_scales}, smaller neighborhoods (\(k=5\)) enhance fine-grained details such as object parts, whereas larger neighborhoods (\(k=20\)) capture room-level spatial layouts.  Integrating all scales yields the best results 70.6~AP\(_{25}\) and 51.2~AP\(_{50}\), demonstrating that hierarchical context aggregation is crucial for complex indoor scenes. This progressive multi-scale reasoning enables the detector to more accurately localize spatially correlated objects—such as chairs around tables or monitors on desks—while maintaining robustness to clutter and occlusion.

%-----------------------------------------------------------------------------------------
\subsection{Discussion and Limitations}

Although GraphFusion3D achieves strong multimodal performance on SUN RGB-D, several limitations should be acknowledged. 
First, our point-only variant attains 72.8 AP$_{25}$ and 57.9 AP$_{50}$ on ScanNetV2, which lags behind several specialized point-based methods such as UniDet3D and V-DETR. 
This gap arises because our architecture is primarily designed and optimized for cross-modal fusion. 
It focuses on effectively injecting rich semantic information from images (such as color and texture) into sparse and incomplete point clouds, 
but underutilizes pure geometric correlations when image data is unavailable.

On ScanNetV2, our full multimodal model (P+I) achieves 75.1 AP$_{25}$ and 60.8 AP$_{50}$, representing a clear improvement of +2.3 AP$_{25}$ over the point-only baseline. 
Nevertheless, it does not reach the current state-of-the-art performance on this benchmark. 
The main reason is that we used only 10 multi-view images per scan during training, significantly fewer than the more than 200 views available in the dataset. 
This limited visual coverage prevents the model from fully exploiting complementary RGB cues across large and complex indoor scenes. 
In contrast, on SUN RGB-D, which provides a single dense RGB-D image per scene, our adaptive cross-modal design delivers more substantial gains, 
achieving state-of-the-art results of 70.6 AP$_{25}$ and 51.2 AP$_{50}$. 
These results demonstrate that GraphFusion3D is particularly effective when rich image features are reliably available to compensate for point-cloud sparsity, occlusion, and missing geometric details.

Furthermore, the inclusion of the image encoder and cross-modal attention mechanisms in the Adaptive Cross-Modal Transformer (ACMT) introduces a moderate computational overhead. 
As reported in Table 4, adding ACMT reduces the inference speed from 16.39 FPS (baseline) to 10.01 FPS for the full model on SUN RGB-D. 
While this trade-off is justified by the accuracy improvements in many practical scenarios, it may constrain deployment in highly resource-constrained or strict real-time applications compared to lightweight point-only detectors.

Despite the above limitations, GraphFusion3D provides clear advantages through its context-aware fusion strategy. 
The Graph Reasoning Module (GRM) explicitly models multi-scale neighborhood relationships among proposals, 
the Cross-Modal Gating (CMG) dynamically balances geometric and semantic cues according to local reliability, 
and the Progressive Cascaded Refinement Decoder iteratively improves localization quality. 
By effectively leveraging rich image features to enhance point representations, the framework delivers robust 3D object detection in challenging indoor environments, 
particularly excelling on single-view RGB-D data while still providing consistent multimodal gains even under limited image coverage. 
Future work will aim to enhance robustness in single-modality settings and further improve computational efficiency for broader applicability.

%----------------------------------------------------------------------------------------
\section{Conclusion}

We presented GraphFusion3D, a unified framework for indoor 3D object detection that effectively fuses geometric and visual information. Our approach integrates three key components: an Adaptive Cross-Modal Transformer (ACMT) for dynamic image-point fusion, a Graph Reasoning Module (GRM) for contextual relationship modeling, and a Cascaded Refinement Decoder for progressive localization. Extensive experiments demonstrate that GraphFusion3D achieves state-of-the-art performance on the SUN RGB-D benchmark and competitive results on ScanNetV2, highlighting the effectiveness of our context-aware fusion and hierarchical reasoning for accurate and robust 3D detection in complex indoor scenes.
{
    \bibliographystyle{ACM-Reference-Format}
    \bibliography{main}
}

\end{document}